\documentclass[10pt,twocolumn,letterpaper]{article}

\usepackage{amsmath,amsfonts,bm}

\def\eqref#1{equation~\ref{#1}}
\def\1{\bm{1}}

\DeclareMathAlphabet{\mathsfit}{\encodingdefault}{\sfdefault}{m}{sl}
\SetMathAlphabet{\mathsfit}{bold}{\encodingdefault}{\sfdefault}{bx}{n}

\DeclareMathOperator*{\argmax}{arg\,max}

\DeclareMathOperator{\sign}{sign}

\usepackage{multirow}
\usepackage{url}
\usepackage{graphicx}
\usepackage{subcaption}
\usepackage{booktabs}
\usepackage{cvpr}
\usepackage{times}
\usepackage{epsfig}
\usepackage{graphicx}
\usepackage{amsmath}
\usepackage{amssymb}
\usepackage{authblk}
\usepackage{multibib}

\usepackage[pagebackref=true,breaklinks=true,letterpaper=true,colorlinks,bookmarks=false]{hyperref}

\cvprfinalcopy % *** Uncomment this line for the final submission

\def\cvprPaperID{2319} % *** Enter the CVPR Paper ID here

\ifcvprfinal\fi
\begin{document}

\definecolor{orange}{RGB}{255,127,0}
\newif\ifnote
\notetrue
\newcommand{\ADnote}[1]{\ifnote \textcolor{blue}{[{\em {\bf **Abhi:} #1}]} \fi}
\newcommand{\DMnote}[1]{\ifnote \textcolor{red}{[{\em {\bf **Dhruv:} #1}]} \fi}
\newcommand{\LVDMnote}[1]{\ifnote \textcolor{green}{[{\em {\bf **Laurens:} #1}]} \fi}
\newcommand{\ZYnote}[1]{\ifnote \textcolor{magenta}{[{\em {\bf **Zeki:} #1}]} \fi}
\newcommand{\YLnote}[1]{\ifnote \textcolor{orange}{[{\em {\bf **Yixuan:} #1}]} \fi}

\newcommand{\x}[0]{\mathbf{x}}
\newcommand{\s}[0]{\mathbf{s}}
\title{Defense Against Adversarial Images using Web-Scale Nearest-Neighbor Search}

% Authors must not appear in the submitted version. They should be hidden
% as long as the \iclrfinalcopy macro remains commented out below.
% Non-anonymous submissions will be rejected without review.

\author[1, 2]{Abhimanyu Dubey\thanks{This work was done while Abhimanyu Dubey was at Facebook AI.}}
\author[2]{Laurens van der Maaten}
\author[2]{Zeki Yalniz}
\author[2]{Yixuan Li}
\author[2]{Dhruv Mahajan}

\affil[1]{Massachusetts Institute of Technology}
\affil[2]{Facebook AI}

% \author{Abhimanyu Dubey} \\
% MIT\\
% \texttt{dubeya@mit.edu} \\
% \And
% Laurens van der Maaten \& Zeki Yalniz \& Yixuan Li \& Dhruv Mahajan \\
% Facebook AI \\
% \texttt{\{lvdmaaten,izy,yixuanl,dhruvm\}@fb.com}
% }

% The \author macro works with any number of authors. There are two commands
% used to separate the names and addresses of multiple authors: \And and \AND.
%
% Using \And between authors leaves it to \LaTeX{} to determine where to break
% the lines. Using \AND forces a linebreak at that point. So, if \LaTeX{}
% puts 3 of 4 authors names on the first line, and the last on the second
% line, try using \AND instead of \And before the third author name.

\newcommand{\fix}{\marginpar{FIX}}
\newcommand{\new}{\marginpar{NEW}}
\newcommand{\comment}[1]{}  %comment not showed

% \iclrfinalcopy % Uncomment f

\maketitle

\begin{abstract}
A plethora of recent work has shown that convolutional networks are not robust to \emph{adversarial images}: images that are created by perturbing a sample from the data distribution as to maximize the loss on the perturbed example. In this work, we hypothesize that adversarial perturbations move the image away from the image manifold in the sense that there exists no physical process that could have produced the adversarial image. This hypothesis suggests that a successful defense mechanism against adversarial images should aim to project the images back onto the image manifold. We study such defense mechanisms, which approximate the projection onto the unknown image manifold by a nearest-neighbor search against a web-scale image database containing tens of billions of images. Empirical evaluations of this defense strategy on ImageNet suggest that it is very effective in attack settings in which the adversary does not have access to the image database. We also propose two novel attack methods to break nearest-neighbor defenses, and demonstrate conditions under which nearest-neighbor defense fails. We perform a series of ablation experiments, which suggest that there is a trade-off between robustness and accuracy in our defenses, that a large image database (with hundreds of millions of images) is crucial to get good performance, and that careful construction the image database is important to be robust against attacks tailored to circumvent our defenses.

\end{abstract}

\section{Introduction}
\begin{figure*}[t]
\centering
\small
\centering
\includegraphics[width=0.85\linewidth]{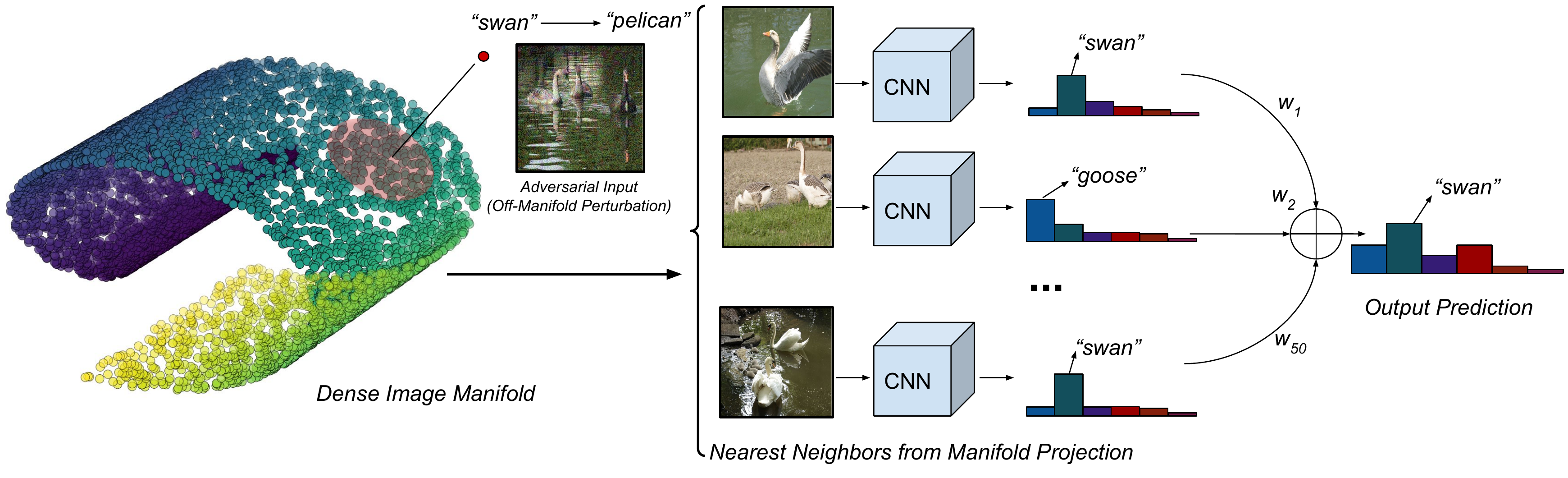}
\caption{Illustration of our defense procedure for improving adversarial robustness in image classification. We first ``project'' the image on to the image manifold by finding the nearest neighbors in the image database, followed by a weighted combination of the predictions of this nearest neighbor set to produce our final prediction.}
\label{fig:teaser}
\end{figure*}
\label{sec:intro}
A range of recent studies has demonstrated that many modern machine-learning models are not robust to \textit{adversarial examples}: examples that are intentionally designed to be misclassified by the models, whilst being nearly indistinguishable from regular examples in terms of some distance measure. Whilst adversarial examples have been constructed against speech recognition \cite{carlini2018audio} and text classification \cite{ebrahimi2018hotflip} systems, most recent work on creating adversarial examples has focused on computer vision~\cite{carlini2017towards, goodfellow2014explaining, kurakin2016adversarial, madry2017towards, moosavi2016deepfool, szegedy2013intriguing}, in which adversarial images are often perceptually indistinguishable from real images. Such adversarial images have successfully fooled systems for image classification~\cite{szegedy2013intriguing}, object detection~\cite{DBLP:journals/corr/XieWZZXY17}, and semantic segmentation~\cite{fischer2017adversarial}. In practice, adversarial images are constructed by maximizing the loss of the machine-learning model (such as a convolutional network) with respect to the input image, starting from a ``clean'' image. This maximization creates an adversarial perturbation of the original image; the perturbation is generally constrained or regularized to have a small $\ell_p$-norm in order for the adversarial image to be perceptually (nearly) indistinguishable from the original.

Because of the way they are constructed, many adversarial images are different from natural images in that there exists no physical process by which the images could have been generated. Hence, if we view the set of all possible natural images\footnote{For simplicity, we ignore synthetic images such as drawings.} as samples from a manifold that is embedded in image space, many adversarial perturbations may be considered as transformations that take a sample from the image manifold and move it away from that manifold. This hypothesis suggests an obvious approach for implementing \emph{defenses} that aim to increase the robustness of machine-learning models against ``off-manifold'' adversarial images \cite{papernot2018deepk, zhao2018retrieval}: \emph{viz.}, projecting the adversarial images onto the image manifold before using them as input into the model.

As the true image manifold is unknown, in this paper, we develop defenses that approximate the image manifold using a massive database of tens of billions of web images. Specifically, we approximate the projection of an adversarial example onto the image manifold by the finding nearest neighbors in the image database. Next, we classify the ``projection'' of the adversarial example, \emph{i.e.}, the identified nearest neighbor(s), rather than the adversarial example itself. Using modern techniques for distributed approximate nearest-neighbor search to make this strategy practical, we demonstrate the potential of our approach in ImageNet classification experiments. Our contributions are:
\begin{enumerate}
    \item We demonstrate the feasibility of web-scale nearest-neighbor search as a defense mechanism against a variety of adversarial attacks in both \textit{gray-box} and \textit{black-box} attack settings, on an image database of an unprecedented scale ($\sim50$ billion images). We achieve robustness comparable to prior state-of-the-art techniques in gray-box and black-box attack settings in which the adversary is unaware of the defense strategy.
    \item To analyze the performance of our defenses in \emph{white-box} settings in which the adversary has full knowledge of the defense technique used, we develop two novel attack strategies designed to break our nearest-neighbor defenses. Our experiments with these attacks show that our defenses break in pure white-box settings, but remain effective in attack settings in which the adversary has access to a comparatively small image database and the defense uses a web-scale image database, even when architecture and model parameters are available to the adversary.
\end{enumerate}
We also conduct a range of ablation studies, which show that: (1) nearest-neighbor predictions based on earlier layers in a convolutional network are more robust to adversarial attacks and (2) the way in which the image database for nearest-neighbor search is constructed substantially influences the robustness of the resulting defense.

\section{Related Work}
After the initial discovery of adversarial examples~\cite{szegedy2013intriguing}, several adversarial attacks have been proposed that can change model predictions by altering the image using a perturbation with small $\ell_2$ or $\ell_{\infty}$ norm~\cite{carlini2017towards, goodfellow2014explaining, kurakin2016adversarial, madry2017towards, moosavi2016deepfool}. In particular,~\cite{madry2017towards} proposed a general formulation of the fast gradient-sign method based on projected gradient descent (PGD), which is currently considered the strongest attack. %While we show the effectiveness of our approach against existing strong attacks, we also propose two novel attack settings designed specifically to break the nearest-neighbor defenses. To the best of our knowledge, nearest-neighbor attack mechanisms have not been explored previously.

A variety of defense techniques have been studied that aim to increase adversarial robustness\footnote{See \url{https://www.robust-ml.org/defenses/} for details.}. \textit{Adversarial training}~\cite{goodfellow2014explaining, huang2015learning, kannan2018adversarial, kurakin2016adversarial} refers to techniques that train the network with adversarial examples added to the training set. Defensive distillation~\cite{DBLP:journals/corr/PapernotMWJS15, papernot2017extending} tries to increase robustness to adversarial attacks by training models using model distillation. 
Input-transformation defenses try to remove adversarial perturbations from input images via JPEG compression, total variation minimization, or image quilting~\cite{dziugaite2016jpg,guo2017countering}. Certifiable defense approaches~\cite{sinha2017certifiable, raghunathan2018certified} aim to guarantee robustness under particular attack settings. Other studies have used out-of-distribution detection approaches to detect adversarial examples~\cite{lee2018simple}. Akin to our approach, PixelDefend~\cite{song2017pixeldefend} and Defense-GAN~\cite{samangouei2018defensegan} project adversarial images back onto the image manifold, but they do so using parametric density models rather than a non-parametric one.

Our work is most closely related to the nearest-neighbor defenses of~\cite{papernot2018deepk,zhao2018retrieval}. \cite{zhao2018retrieval} augments the convolutional network with an off-the-shelf image retrieval system to mitigate the adverse effect of ``off-manifold'' adversarial examples, and uses local mixup to increase robustness to ``on-manifold'' adversarial examples. In particular, inputs are projected onto the feature-space convex hull formed by the retrieved neighbors using trainable projection weights; the feature-producing convolutional network and the projection weights are trained jointly. In contrast to~\cite{zhao2018retrieval}, our approach does not involve alternative training procedures and we do not treat on-manifold adversarial images separately~\cite{gilmer2018advsphere}.

%By leveraging billions of web images for nearest-neighbor retrieval, the effect of on-manifold adversarial examples is implicitly reduced.  \footnote{\cite{gilmer2018advsphere} also mention that it is not conclusive that their observations about off-manifold adversarial examples clearly apply to real-world datasets.}.

\comment{
After the discovery of adversarial examples in deep neural networks in the work of~\cite{szegedy2013intriguing}, there have been several strands of research in the area of adversarial examples, in terms of both defense techniques and adversarial sample generation. The work of~\cite{goodfellow2014explaining} introduced the first reliable method for generating adversarial samples, which was extended to the iterative setting in~\cite{kurakin2016adversarial}. \cite{carlini2017towards} introduce an alternative method of generating adversarial samples by minimizing a convex surrogate for the top-predicted class probability directly. DeepFool~\cite{moosavi2016deepfool} iteratively solves the closed-form version problem of finding the minimal perturbation (under the $\ell_2$-norm) that changes the initially predicted class. More recently,~\cite{madry2017towards} proposed a stronger, more general formulation of the gradient-sign method via projected gradient descent (PGD).

A variety of different defense techniques have been proposed to handle adversarial samples. \textit{Adversarial training}~\cite{goodfellow2014explaining, huang2015learning} encompasses a broad class of defense techniques that involve retraining the network with adversarial examples added to the training set. While this baseline approach is effective, it has been shown to be ineffective against stronger adversaries~\cite{carlini2017towards}. Within the adversarial training regime, several defenses have been proposed with different training routines.~\cite{kurakin2016adversarial} demonstrated the feasibility of adversarial training for large-scale problems such as ImageNet training. A stronger objective that matches the individual logits between adversarial and clean images was introduced in~\cite{kannan2018adversarial}. For smaller problems,~\textit{certifiably} robust adversarial training methods have also been introduced in~\cite{sinha2017certifiable, raghunathan2018certified}. \cite{DBLP:journals/corr/PapernotMWJS15} introduce \textit{Defensive distillation}, a defense that involves training a network by knowledge transfer from a separate teacher network, which was extended in~\cite{papernot2017extending}, by modifying the labelling information used by the distillation technique in handling adversarial samples.}

\section{Problem Setup}
\begin{figure*}[t]
\centering
\small
\centering
\includegraphics[width=0.85\linewidth]{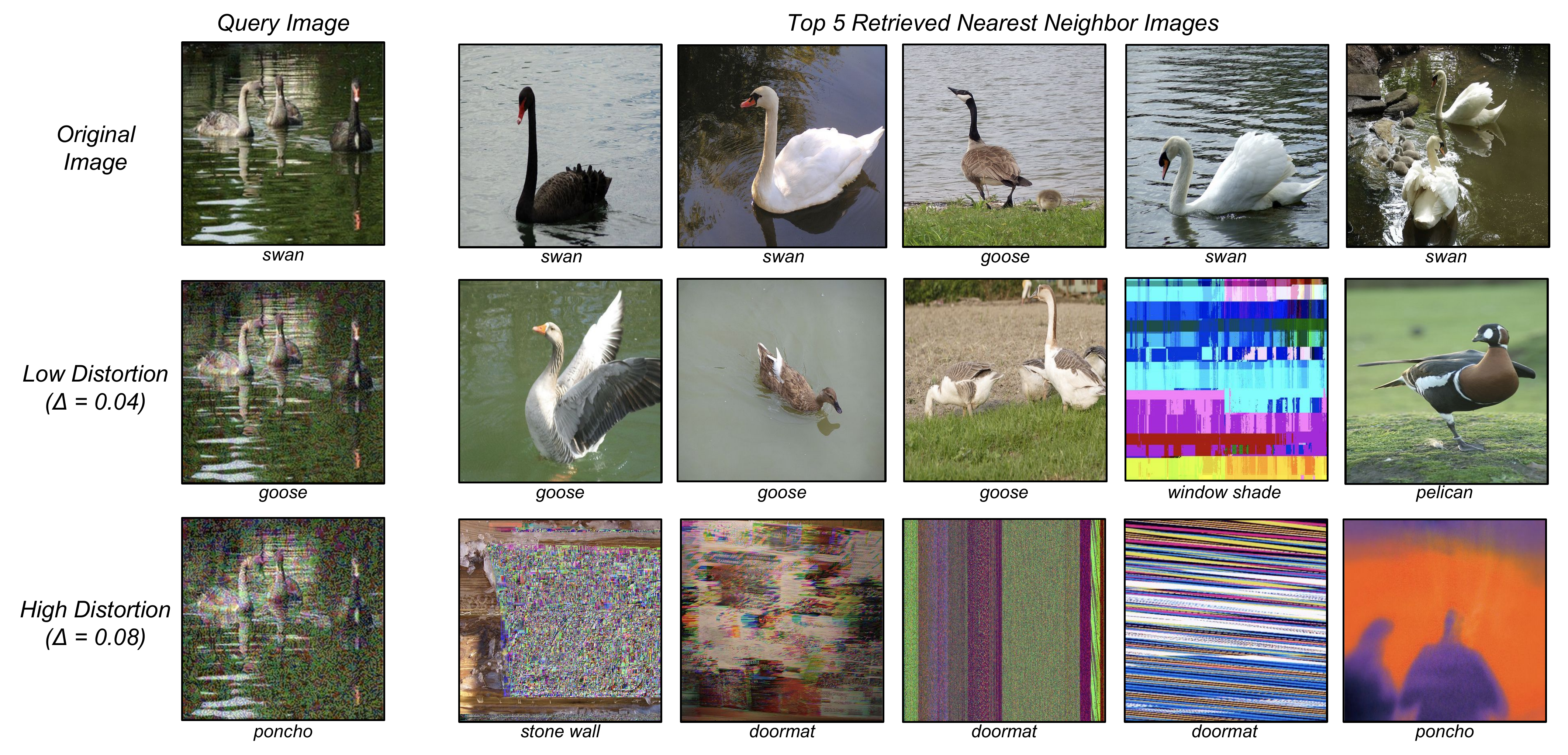}
\caption{Visualization of an image and its five nearest neighbors in the \textit{YFCC-100M} database (from left to right) based on \texttt{conv\_5\_1} features for a clean image (top), an image with a small adversarial perturbation ($\Delta\!=\!0.04$; center), and an image with a large adversarial perturbation ($\Delta\!=\!0.08$; bottom). Adversarial images generated using PGD with a ResNet-50 trained on ImageNet.}
\label{fig:visualization}
\end{figure*}
% \LVDMnote{Please be consistent in the notation here. A mathcal letter cannot represent a distribution, a set, and space. (I think both $\mathcal D$ and $\mathcal H$ are sets.) Use only lower-case or upper-case for scalar constants (I prefer upper-case); the two are mixed now. Be consistent about boldfacing vectors.}
We consider multi-class image classification of images $\x \in [0, 1]^{H \times W}$ into one of $C$ classes. We assume we are given a labeled training set with $N$ examples, $\mathcal D = \left\{(\x_1, y_1), ..., (\x_N, y_N) \right\}$ with labels $y \in \mathbb Z_C$. Training a classification model amounts to selecting a hypothesis $h(\x) \rightarrow \mathbb Z_C$ from some hypothesis set $\mathcal H$. The hypothesis set ${\mathcal H}$ is the set of all possible parameter values for a convolutional network architecture (such as a ResNet), and the hypothesis $h(\x)$ is selected using empirical risk minimization: specifically, we minimize the sum of a loss function $L(\x_n, y_n; h)$ over all examples in $\mathcal D$ (we omit $h$ where it is obvious from the context). Throughout the paper, we choose $L(\cdot, \cdot;\cdot)$ to be the multi-class logistic loss.
% , \emph{i.e.}, the cross entropy between the prediction $h(\x_n)$ and the corresponding label $y_n$.

\subsection{Attack Model}
Given the selected hypothesis (\emph{i.e.}, the model) $h \in \mathcal H$, the adversary aims to find an adversarial version $\x^*$ of a real example $\x$ for which: (1) $\x^*$ is similar to $\x$ under some distance measure and (2) the loss $L(h(\x^*), y)$ is large, \emph{i.e.}, the example $\x^*$ is likely to be misclassified. In this paper, we measure similarity between $\x^*$ and $\x$ by the normalized $\ell_2$ distance\footnote{Other choices for measuring similarity include the $\ell_\infty$ metric~\cite{WardeFarley20161AP}.}, given by $\Delta(\x, \x^*)\!=\!\frac{\lVert \x - \x^* \rVert_2}{\lVert\x  \rVert_2}$. Hence, the adversary's goal is to find for some similarity threshold $\epsilon$:
\begin{equation*}
    \x^*  = \argmax_{\x' : \Delta(\x, \x') \leq \epsilon} L(\x', y; h).
\end{equation*}
Adversarial attacks can be separated into three categories: (1) \textit{white-box attacks}, where the adversary has access to both the model $h$ and the defense mechanism; (2) \textit{black-box attacks}, where the adversary has no access to $h$ nor the defense mechanism; and (3) \textit{gray-box attacks} in which the adversary has no direct access to $h$ but has partial information of the components that went into the construction of $h$, such as the training data $\mathcal D$, the hypothesis set $\mathcal H$, or a superset of the hypothesis set $\mathcal H$. While robustness against \textit{white-box} adversarial attacks is desirable since it is the strongest notion of security~\cite{Papernot2017practicalbb}, in real-world settings, we are often interested in robustness against \textit{gray-box} attacks because it is rare for an adversary to have complete information (cf. \textit{white-box}) or no information whatsoever (cf. \textit{black-box}) on the model it is attacking~\cite{ilyas2018bb}.
% \LVDMnote{There was a paper recently studying attacks under a budgeted query-access setting. We should refer to it here.}

\subsection{Adversarial Attack Methods}
\label{sec:attackmethods}
%First, we experiment with different adversarial attack methods that do not exploit the nearest-neighbor nature of our defense.
The \textbf{Iterative Fast Gradient Sign Method (I-FGSM)}~\cite{kurakin2016adversarial} generates adversarial examples by iteratively applying the following update for $m\!=\!\{1, ..., M\}$ steps:
\begin{multline*}
        \x^{(m)} = \x^{(m-1)} + \varepsilon \cdot \sign \left(\nabla_{\x^{(m-1)}} L(\x^{(m-1)}, y)\right), \\ \text{where } \x^*_{\textsf{IFGSM}} = \x^{(M)}, \ \text{and } \x^{(0)} = \x.
\end{multline*}
When the model is available to the attacker (\textit{white-box} setting), the attack can be run using the true gradient $\nabla_\x  L(h(\x), y)$, however, in \textit{gray-box} and \textit{black-box} settings, the attacker has access to a surrogate gradient $\nabla_\x  L(h'(\x), y)$, which in practice, has been shown to be effective as well. The \textbf{Projected Gradient Descent (PGD)}~\cite{madry2017towards} attack generalizes the I-FGSM attack by: (1) clipping the gradients to project them on the constraints formed by the similarity threshold and (2) including random restarts in the optimization process. Throughout the paper, we employ the PGD attack in our experiments because recent benchmark competitions suggest it is currently the strongest attack method.

In the appendix, we also show results with \textbf{Fast Gradient Sign Method (FGSM)}~\cite{goodfellow2014explaining} and \textbf{Carlini-Wagner's $\ell_p$ (CW-Lp)}~\cite{carlini2017towards} attack methods.
 For all the attack methods, we use the implementation of~\cite{guo2017countering} and enforce that the image remains within $[0,1]^{H\times W}$ by clipping pixel values to lie between $0$ and $1$.

\section{Adversarial Defenses via Nearest Neighbors}
\label{sec:knndesc}
The underlying assumption of our defense is that adversarial perturbations move the input image away from the image manifold. The goal of our defense is to project the images back onto the image manifold before classifying them. As the true image manifold is unknown, we use a sample approximation comprising a database of billions of natural images. When constructing this database, the images may be selected in a weakly-supervised fashion to match the target task, for instance, by including only images that are associated with labels or hashtags that are relevant to that task~\cite{mahajan2018exploring}. To ``project'' an image on the image manifold, we identify its $K$ nearest neighbors from the database by measuring Euclidean distances in some feature space. Modern implementations of approximate nearest neighbor search allow us to do this in milliseconds even when the database contains billions of images \cite{johnson2017billion}. Next, we classify the ``projected'' adversarial example by classifying its nearest neighbors using our classification model and combining the resulting predictions. In practice, we pre-compute the classifications for all images in the image database and store them in a key-value map \cite{ghemawat2011leveldb} to make prediction efficient.

We combine predictions by taking a weighted average of the \emph{softmax probability vector} of all the nearest neighbors\footnote{In preliminary experiments, we also tried averaging ``hard'' rather than ``soft'' predictions but we did not find that to work better in practice.}. The final class prediction is the $\arg\max$ of this average vector. We study three strategies for weighting the importance of each of the $K$ predictions in the overall average:%This corresponds to a ``soft'' combination of the model predictions over the data manifold.\vspace*{0.05in}

\begin{table*}[h]
\centering
\setlength\tabcolsep{3pt}
\begin{tabular}{l|ccc|ccc|ccc}
\toprule
\multirow{2}{*}[-0.6em]{\textbf{Image database}} & \multicolumn{3}{c|}{\textbf{Clean}} & \multicolumn{3}{c|}{\textbf{Gray box}} & \multicolumn{3}{c}{\textbf{Black box}} \\
& UW & CBW-E & CBW-D & UW & CBW-E & CBW-D & UW & CBW-E & CBW-D\\ \midrule
IG-50B-All (\texttt{conv\_5\_1-RMAC}) & 0.632 & 0.644 & {\bf 0.676} & 0.395 & 0.411 & {\bf 0.427} & 0.448  & 0.459 & {\bf 0.491} \\
IG-1B-Targeted (\texttt{conv\_5\_1}) & 0.659 & 0.664 & {\bf 0.681} & 0.415 & 0.429 & {\bf 0.462} & 0.568 & 0.574 & {\bf 0.587} \\
IN-1.3M (\texttt{conv\_5\_1}) & 0.472 & 0.469 & 0.471 & 0.285 & 0.286 & 0.286 & 0.311 & 0.312 & 0.312 \\
\bottomrule
\end{tabular}
\caption{ImageNet classification accuracies of a ResNet-50 using our nearest-neighbor defense with three different weighting strategies (UW, CBW-E, and CBW-D) on PGD adversarial ImageNet images with a normalized $\ell_2$ distance of $0.06$. Nearest-neighbor searches were performed on three image databases (rows) with $K\!=\!50$. Accuracies using KNN defense on clean images are included for reference.}
\label{tab:apvscbv}
\end{table*}

\textbf{Uniform weighting (UW)} assigns the same weight ($w\!=\!1/K$) to each of the predictions in the average. 

We also experimented with two confidence-based weighting schemes that take into account the ``confidence'' that the classification model has in its prediction for a particular neighbor. This is important because, empirically, we observe that ``spurious'' neighbors exist that do not correspond to any of the classes under consideration, as displayed in Figure~\ref{fig:visualization} (center row, fourth retrieved image). The entropy of the softmax distribution for such neighbors is very high, suggesting we should reduce their contribution to the overall average. We study two measures for compute the weight, $w$, associated with a neighbor: (1) an entropy-based measure, CBW-E(ntropy); and (2) a measure for diversity among the top-scoring classes, CBW-D(iversity).

\textbf{CBW-E} measures the entropy gap between a class prediction and the entropy of a uniform prediction. Hence, for softmax vector $\mathbf s$ over $C$ classes ($\forall c \in \{1, \dots, C\}: s_c \in [0, 1]$ and $\sum_{c \in \{1, \dots, C\}} s_c = 1$), the weight $w$ is given by:
\begin{equation*}
w = \left\lvert\log C + \sum_{c=1}^C s_c \log s_c\right\rvert.
\end{equation*}

$\textbf{CBW-D}$ computes $w$ as a function of the difference between the maximum value of the softmax distribution and the next top $M$ values. Specifically, let $\mathbf{\hat{s}}$ be the sorted (in descending order) version of the softmax vector $\mathbf{s}$. The weight $w$ is defined as:
\begin{equation*}
w = \sum_{m=2}^{M+1} (\hat{s}_1 - \hat{s}_m)^P.
\end{equation*}
We tuned $M$ and $P$ using cross-validation in preliminary experiments, and set $M \!=\! 20$ and $P \!=\! 3$ in all experiments that we present in the paper.

\section{Experiments: Gray and Black-Box Settings}
\label{sec:Experiments}
To evaluate the effectiveness of our defense strategy, we performed a series of image-classification experiments on the ImageNet dataset. Following~\cite{kannan2018adversarial}, we assume an adversary that uses the state-of-the-art PGD adversarial attack method (see Section~\ref{sec:attackmethods}) with $10$ iterations. In the appendix, we also present results obtained using other attack methods.
%In this section, we perform experiments demonstrating the performance of different nearest-neighbor settings discussed in Section~\ref{sec:knndesc} against adversarial perturbations and also compare with the existing defense methods in the literature. In Section~\ref{sec:apvscbv}, we investigate different ways of combining the predictions. Section~\ref{sec:knnfeat} shows the robustness vs. accuracy trade-offs incurred by features at different depths of neural network. In Section~\ref{sec:knnsize}, we study the effect of size of index on the performance. Section~\ref{sec:knnlabel} investigates how different label distributions of the images in index set impacts the KNN defense. Finally, in Section~\ref{sec:soacomp}, we compare our method against existing strong defense methods showing increased robustness against adversarial attacks.

\subsection{Experimental Setup}
\label{sec:Experimental Setup}
To perform image classification, we use ResNet-18 and ResNet-50 models \cite{he2016deep} that were trained on the ImageNet training set. We consider two different attack settings: (1) a \emph{gray-box attack} setting in which the model used to generate the adversarial images is the same as the image-classification model, \emph{viz.} the ResNet-50; and (2) a \emph{black-box attack} setting in which the adversarial images are generated using the ResNet-18 model and the prediction model is ResNet-50 (following \cite{guo2017countering}). We experiment with a number of different implementations of the nearest-neighbor search defense strategy by varying: (1) the image database that is queried by the defense and (2) the features that are used as basis for the nearest-neighbor search.

\textbf{Image database.} We experiment with three different web-scale image databases as the basis for our nearest-neighbor defense.
\begin{itemize}
\item \textit{IG-$N$-$\star$} refers to a database of $N$ public images with associated hashtags that is collected from a social media website, where $\star$ can take two different values. Specifically, \textit{IG-$N$-All} comprises images that were selected at random. Following~\cite{mahajan2018exploring}, {\it{IG-$N$-Targeted}} contains exclusively images that were tagged with at least one $1,500$ hashtags that match one of the $1,000$ classes in the ImageNet-1K benchmark. The $1,500$ hashtags were obtained by canonicalizing all synonyms corresponding to the synsets, which is why the dataset contains more hashtags than classes. Our largest database contains $N\!=$ 50 billion images.
\item \textit{YFCC-100M} is a publicly available dataset of 100 million Flickr images with associated meta-data~\cite{thomee2015yfcc}.
\item \textit{IN-1.3M} refers to the training split of the publicly ImageNet-1K dataset of approximately 1.28M images.
\end{itemize}

\textbf{Features.} We constructed feature representations for the images in each of these image databases were constructed by: (1) computing pre-ReLU activations from the \texttt{conv\_2\_3}, \texttt{conv\_3\_4}, \texttt{conv\_4\_6}, or \texttt{conv\_5\_1} layer of a ResNet-50 trained on ImageNet-1K and (2) reducing these feature representations to 256 dimensions using a spatial average pooling followed by PCA. For our largest database of 50 billion images, we used a feature representation that requires less storage: following~\cite{mahajan2018exploring}, we use \texttt{conv\_5\_1-RMAC} features that were obtained by using \texttt{conv\_5\_1} features from a ResNet-50 model followed by R-MAC pooling~\cite{tolias2015particular}, bit quantization, and dimensionality reduction (see the appendix for details).

\subsection{Results}
\label{sec:apvscbv}
Table~\ref{tab:apvscbv} presents the classification accuracy of a ResNet-50 using our defense strategy on PGD adversarial ImageNet images with a normalized $\ell_2$ dissimilarity of $0.06$; the table presents results in both the gray-box and the black-box settings. The table presents results for both the uniform weighting (UW) and the confidence-based weighting (CBW-E and CBW-D) strategies, performing nearest-neighbor search in three different image databases using  a value of $K\!=\!50$. The results presented in the table demonstrate the potential of large-scale nearest neighbors as a defense strategy: our best model achieves a top-1 accuracy of $46.2\%$ in the gray-box and of $58.7\%$ in the black-box setting. The results also show that CBW-D weighting consistently outperforms the other weighting schemes, and that the use of web-scale IG-$N$-* databases with billions of images leads to substantially more effective defenses than using the ImageNet training set. %In the hard combination case, we observe that CBW (both CBW-D and CBW-E) do not provide major improvements. We proceed with the soft combination and CBW-D weighing scheme for the remainder of our experiments.

\begin{figure}[t]
\centering
\small
\centering
\includegraphics[width=0.75\linewidth]{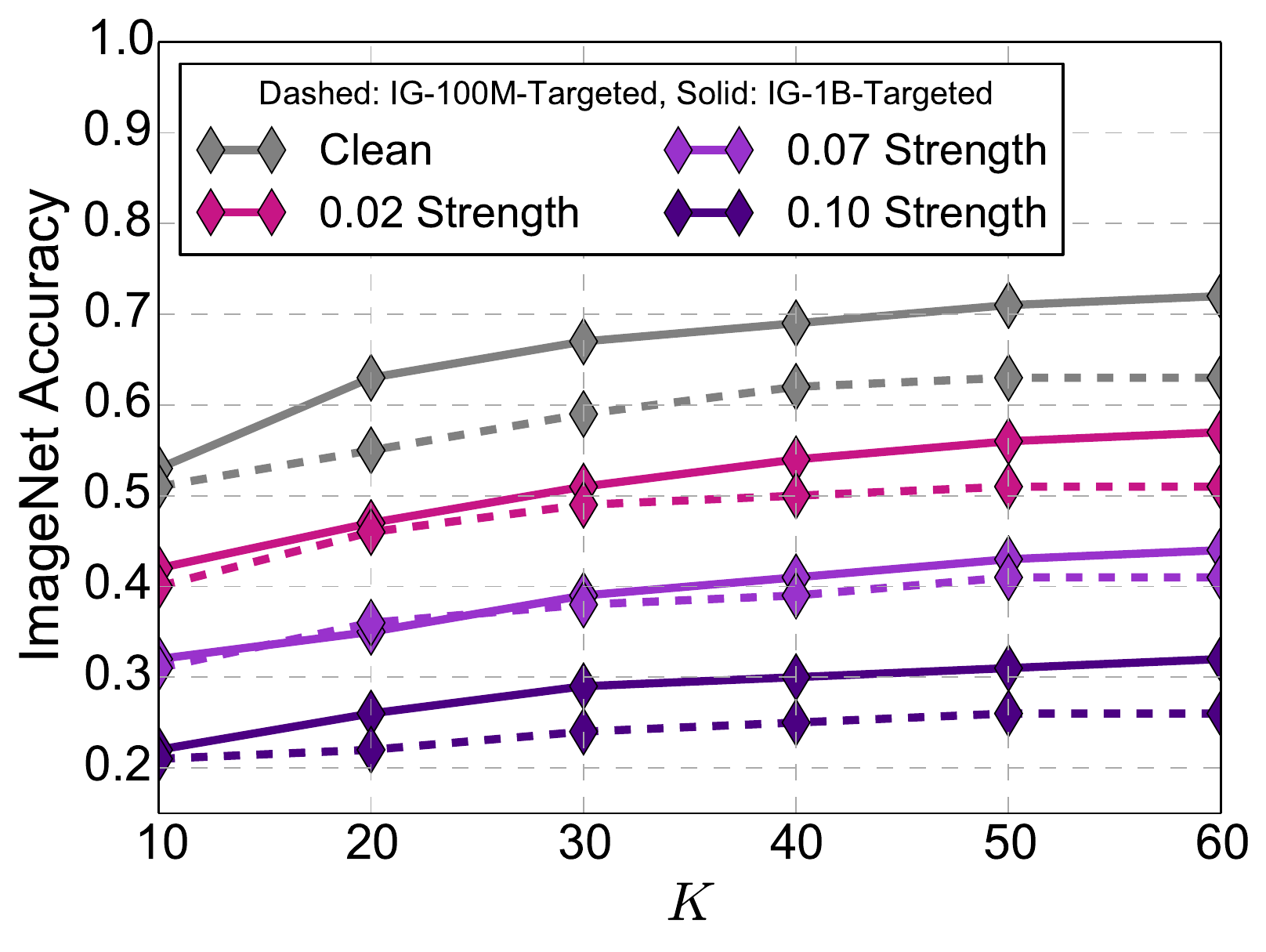}
\caption{Classification accuracy of ResNet-50 using our CBW-D defense on PGD adversarial ImageNet images, as a function of the normalized $\ell_2$ norm of the adversarial perturbation. Defenses are implemented via nearest-neighbor search using \texttt{conv\_5\_1} features on the IG-1B-Targeted (solid lines) and IG-100M-Targeted (dashed lines). Results are for the black-box setting.}
\label{fig:k_variation}
\end{figure}

Motivated by its strong performance in our initial experiments, we use the CBW-D strategy for all the ablation studies and analyses that we perform below.

%To better understand the impact of various design choices on the effectiveness of our defense, we present a series of more detailed ablation studies and analyses below.

\paragraph{How does the choice of $K$ influence the effectiveness of the defense?}
\label{sec:knn_k}
The robustness\footnote{Recent work also proposed a nearest-neighbors algorithm that is claimed to be robust under adversarial perturbations~\cite{wang2017analyzing}. We do not study that algorithm here because it does not scale to web-scale image datasets.} of nearest-neighbor algorithms depends critically on the number of nearest neighbors, $K$~\cite{wang2017analyzing}. Motivated by this observation, we empirically analyze the effect of $K$ on the effectiveness of our defense. Figure~\ref{fig:k_variation} presents the classification accuracy of a ResNet-50 with CBW-D using \texttt{conv\_5\_1} features as a function $K$; results are presented on two different databases, \emph{viz.}, IG-100M-Targeted (dashed lines) and IG-1B-Targeted (solid lines). The results in the figure reveal increasing $K$ appears to have a positive effect on classification accuracy for both databases, although the accuracy appears to saturate at $K \!=\! 50$. Therefore, we use $K \!=\! 50$ in the remainder of our experiments. The figure also shows that defenses based on the larger image database (1B images) consistently outperform those based on the smaller database (100M images).

\begin{figure*}[t]
\centering
\small
\centering
\includegraphics[width=0.8\linewidth]{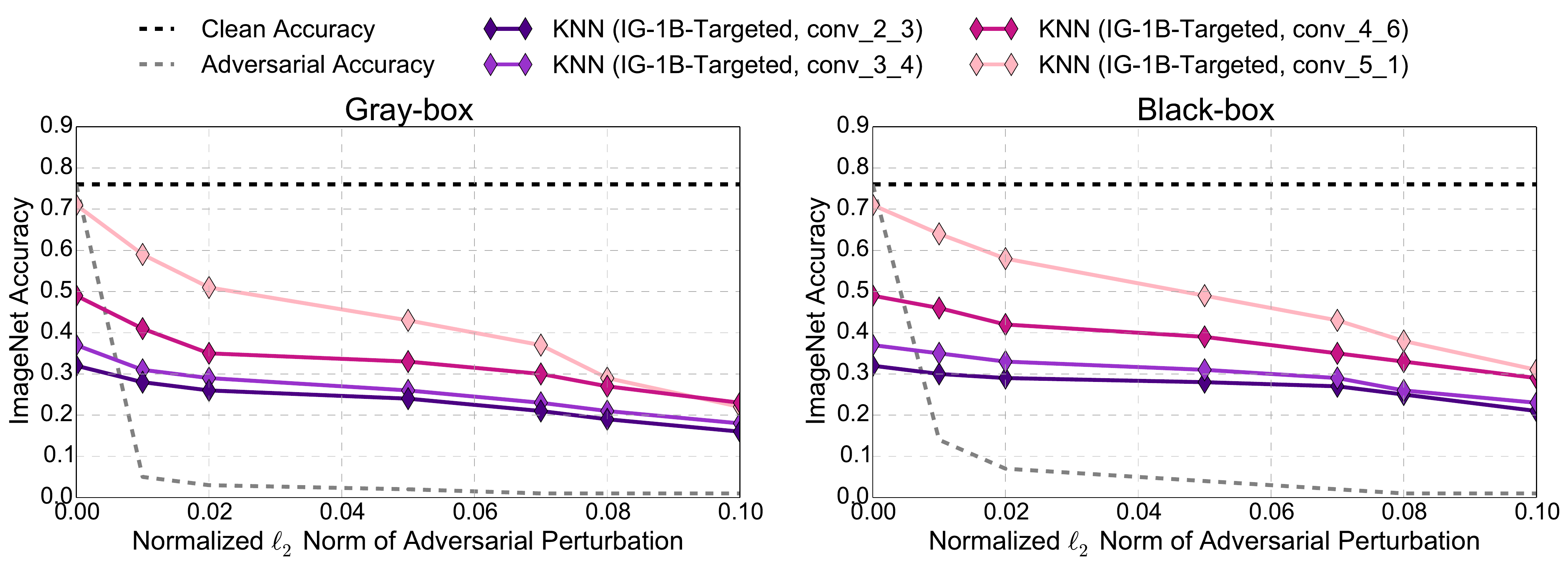}
\caption{Classification accuracy of ResNet-50 using the CBW-D defense on PGD adversarial ImageNet images, as a function of the normalized $\ell_2$ norm of the adversarial perturbation. Defenses use four different feature representations of the images in the IG-1B-Targeted image database. Results are presented for the gray-box (left) and black-box (right) settings.}
\label{fig:feature_variation}
\end{figure*}
\paragraph{How does the choice of features influence the effectiveness of the defense?}
\label{sec:knnfeat}
Figure~\ref{fig:feature_variation} presents classification accuracies obtained using CBW-D defenses based on four different feature representations of the images in the IG-1B-Targeted database. The results presented in the figure show that using ``later'' features for the nearest-neighbor search generally provide better accuracies, but that ``earlier'' features are more robust (\emph{i.e.}, deteriorate less) in the perturbation norm regime we investigated. Earlier features presumably provide better robustness because they are less affected by the adversarial perturbation in the image, which induces only a small perturbation in those features. The downside of using early features is that they are more susceptible to non-semantic variations between the images, which is why the use of later features leads to higher classification accuracies. Motivated by these results, we use \texttt{conv\_5\_1} features in the remainder of our experiments.

\begin{figure*}[t]
\centering
\small
\includegraphics[width=0.8\linewidth]{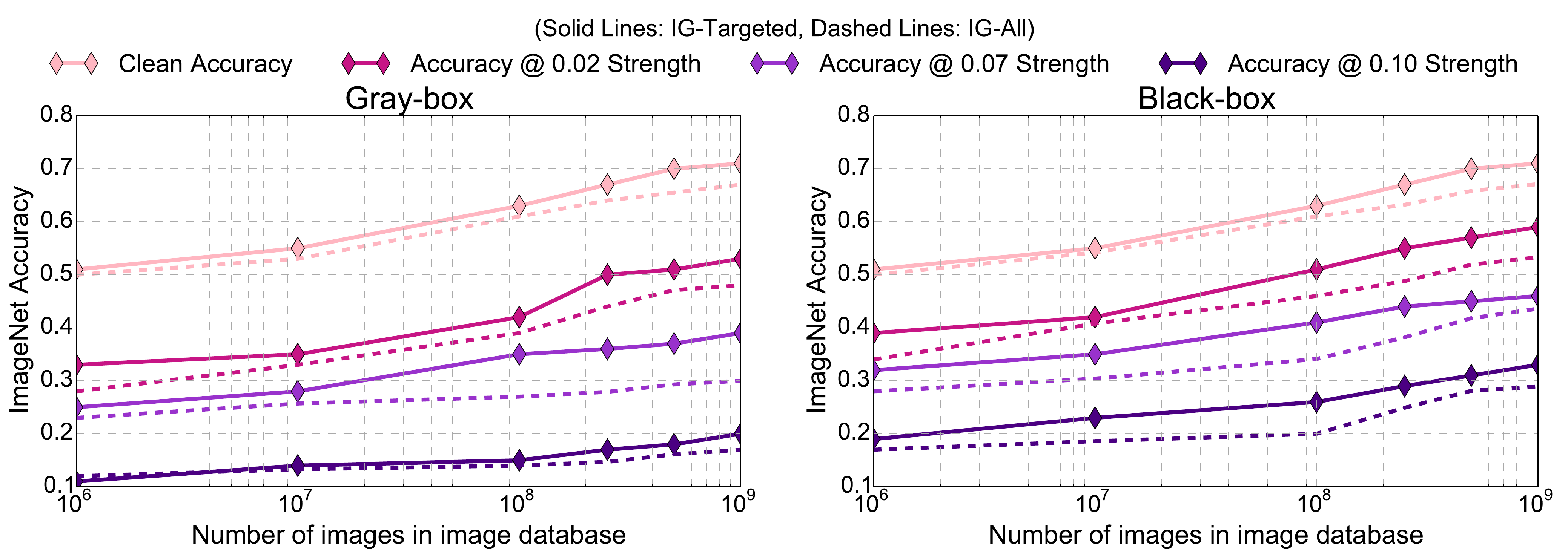}
\caption{Classification accuracy of ResNet-50 using the CBW-D defense on PGD adversarial ImageNet images, using the IG-$N$-Targeted database (solid lines) and IG-$N$-All database (dashed lines) with different values of $N$. Results are presented in the gray-box (left) and black-box (right) settings.}
\label{fig:size_variation}
\end{figure*}

\paragraph{How does nearest-neighbor index size influence the effectiveness of the defense?}
\label{sec:knnsize}
Next, we measured the effectiveness of our defense strategy as a function of the size of the image database used for nearest-neighbor searches. Figure~\ref{fig:size_variation} presents the effect of the image database size ($N$) on the classification accuracy for different attack strengths; experiments were performed on the IG-$N$-Targeted database. In line with earlier work~\cite{mahajan2018exploring,sun2017unreasonable}, the results suggest a log-linear relation between accuracy and database size: each time the database size is doubled, the classification accuracy increases by a fixed number of percentage points. This result appears to be consistent across a range of magnitudes of the adversarial perturbation.

%The results show the need for the index set to be in the regime of at-least hundreds of millions of images to achieve good accuracy; there is a $7$-point improvement in clean accuracy as we increase the size from $10$M to $100$M, and, another $7$-point improvement from $100$M to $1$B. In comparison, robustness improves only marginally with increasing index size. This result is consistent with the recent theoretical results of ~\cite{gilmer2018advsphere}, that a small increase in robustness requires a significant decrease in the generalization error.

\paragraph{How does selection of images in the index influence the effectiveness of the defense?}
\label{sec:knnlabel}
%Constructing a billion-scale index of images to support any arbitrary training set is difficult, since it requires an exhaustive domain-specific labelling of the images in the database. To ameliorate this, one might consider utilizing a general-purpose repository of images.
In Figure~\ref{fig:size_variation}, we also investigate how important it is that the images in the database used for the CBW-D defense are semantically related to the (adversarial) images being classified, by comparing defenses based on the IG-$N$-All and IG-$N$-Targeted databases for varying index sizes. The results reveal that there is a positive effect of ``engineering'' the defense image database to match the task at hand: defenses based on IG-$N$-Targeted outperforms those based on IG-$N$-All by a margin of $1\%\!-\!4\%$ at the same database size.

%The result suggests that while it is important to match the labels in the index set with the target (ImageNet-1K) task, untargeted databases also provide robustness as well as good classification performance.

% In this section, we report results of ablation studies with regard to the type of the database conducted to understand the impact of domain-specificity on adversarial robustness.
% We also find that the IG-ImageNet and ImageNet indexes perform comparably; this indicates that the weakly-supervised technique of building large indexes by approximately matching the labels in the target task is an effective and practical middle path between supervised and un-supervised setting. However, we do not observe a substantial differences in robustness between different types of indexes.

% (i) fully supervised ImageNet1k training data (ImageNet), (ii) weakly-supervised data selected from both ImageNet1K specific tags (IG-ImageNet) and generic set of tags (YFCC), and, (iii) unsupervised data consisting of random images (IG-Random).
%
% \begin{figure*}[t]
% \centering
% \small
% \includegraphics[width=0.9\linewidth]{figures/indextype_combined.pdf}
% \caption{Classification accuracy of ResNet-50 using the CBW-D defense on PGD adversarial ImageNet images, comparing defenses that use IG-$N$-All databases and defenses that use IG-$N$-Targeted databases for two different values of $N$. Results are in the gray-box (left) and black-box (right) attack settings.}
% \label{fig:index_variation}
% \end{figure*}

\subsection{Comparison with State-of-the-Art Defenses}
\label{sec:soacomp}
In Table~\ref{tab:baseline}, we compare the effectiveness of our nearest-neighbor defense with that of other state-of-the-art defense strategies. Specifically, the table displays the classification accuracy on adversarial ImageNet images produced with PGD, using a normalized $\ell_2$ distance of $0.06$. The results in the table show that, despite its simplicity, our defense strategy is at least as effective as alternative approaches (including approaches that require re-training of the network). To the best of our knowledge, our defense strategy even outperforms the current state-of-the-art In the gray-box setting. In the black-box setting, image quilting \cite{guo2017countering} performs slightly better than our nearest-neighbor defense. Interestingly, image quilting is also a nearest-neighbor approach but it operates at the image patch level rather than at the image level.

%We find that using the IG-1B-ImageNet index and Feat-D-50 feature space, we obtain performance that is, to the best of our knowledge, state-of-the-art in the \textit{gray-box} setting. In the \textit{black-box} setting, Image Quilting~\cite{guo2017countering} performs the best, with KNN Defense performing comparably ($\approx 3\%$ difference). Note that Image Quilting is a very expensive operation involving nearest-neighbor search for multiple patches in the image followed by a graph-cut operation, and utilizes the same underlying approach.

\begin{table}[h]
\centering
\small
\resizebox{0.8\linewidth}{!}{
\begin{tabular}{l|ccc}
\toprule
\textbf{Defense} & \textbf{Clean} & \textbf{Gray box} & \textbf{Black box} \\ \midrule
No defense & 0.761 & 0.038 & 0.046 \\ \midrule
Crop ensemble~\cite{guo2017countering} & 0.652 & 0.456 & 0.512 \\
TV Minimization~\cite{guo2017countering} & 0.635 & 0.338 & 0.597 \\
Image quilting~\cite{guo2017countering} & 0.414 & 0.379 & \textbf{0.618} \\
Ensemble training~\cite{tramer2017ensemble} & -- & -- & 0.051 \\
ALP~\cite{kannan2018adversarial} & 0.557 & 0.279 & 0.348 \\
RA-CNN~\cite{zhao2018retrieval}$^*$ & 0.609 & 0.259 & -- \\ \midrule
\multicolumn{4}{c}{\textit{Our Results}} \\ \midrule
IG-50B-All (\texttt{conv\_5\_1-RMAC}) & 0.676 & 0.427 & 0.491 \\
IG-1B-Targeted (\texttt{conv\_5\_1}) & \textbf{0.681} & \textbf{0.462} & 0.587 \\
YFCC-100M (\texttt{conv\_5\_1}) & 0.613 & 0.309 & 0.395 \\
IN-1.3M (\texttt{conv\_5\_1}) & 0.471 & 0.286 & 0.312 \\
\bottomrule
\end{tabular}
}
\caption{ImageNet classification accuracies of ResNet-50 models using state-of-the-art defense strategies against the PGD attack, using a normalized $\ell_2$ distance of $0.06$. $^*$ RA-CNN~\cite{zhao2018retrieval} experiments were performed using a ResNet-18 model.}
\label{tab:baseline}
\end{table}

\section{Experiments: White-Box Setting}
So far, we have benchmarked our adversarial defense technique against PGD attacks that are unaware of our defense strategies, \emph{i.e.}, we have considered gray-box and black-box attack settings. However, in real-world attack settings, we may expect adversaries to be aware of defense mechanism and tailor their attacks to circumvent the defenses. In this section, we study such a \emph{white-box} setting by designing adversarial attacks that try to circumvent nearest-neighbor defenses, and we study the effectiveness of these novel attacks against our defense strategy.

\subsection{Defense-Aware Attacks}
We develop two defense-aware attacks in which the adversary uses nearest-neighbor search on a web-scale image database to simulate the defense. Whilst this ``attack database'' may be identical to the ``defense database'', it is likely that the databases are not exactly the same in practice (if only, because the defender can randomize its defense database). We develop two defense-aware attacks:

\textbf{Nearest-neighbor prediction attack (PGD-PR).} Given an attack database $\widehat{\mathcal{D}}_A$ and a function $g(\x)$ that the attacker uses to compute a feature representation for example $\x$, we first compute the corresponding set of $K$ nearest neighbors, $\widehat{\mathcal{D}}_{A, K}(g(\x))$. Subsequently, we add an extra loss term that maximizes the loss between those $K$ nearest neighbors and the model prediction when constructing the adversarial sample $\x^*$. Specifically, we perform PGD-like updates:
\begin{equation*}
\resizebox{\linewidth}{!}{%
    $\x^* = \x + \varepsilon \cdot \sign \bigg[\nabla_\x L(h(\x), y) + \gamma \sum_{\x' \in \widehat{\mathcal{D}}_{A, K}(\x)} \nabla_{\x'} L(h(\x'), y)\bigg].$%
    }
\end{equation*}
Herein, the hyperparameter $\gamma$ trade off the loss suffered on the sample itself with the loss suffered on its nearest neighbors. We set $\gamma \!=\! 0.05$ and $K \!=\! 50$ in our paper. We perform an iterative PGD-like update using this objective function, where we fix the neighbor set beforehand.

\textbf{Nearest-neighbor feature-space attack (PGD-FS).} In contrast to the previous attack, this attack targets the feature space used for nearest-neighbor retrieval. Specifically, it attacks the feature extractor $g(\x)$ directly by producing the adversarial sample $\x^*$ via PGD-like updates:
\begin{equation*}
\resizebox{0.8\linewidth}{!}{%
    $\x^* = \x + \varepsilon \cdot \sign \left[\sum_{\x' \in \widehat{\mathcal{D}}_{A, K}(\x)} \nabla_{\x'}\lVert g(\x') - g(\x) \rVert_2^2\right].$%
    }
\end{equation*}
\begin{figure*}[t]
\centering
\small
\includegraphics[width=\linewidth]{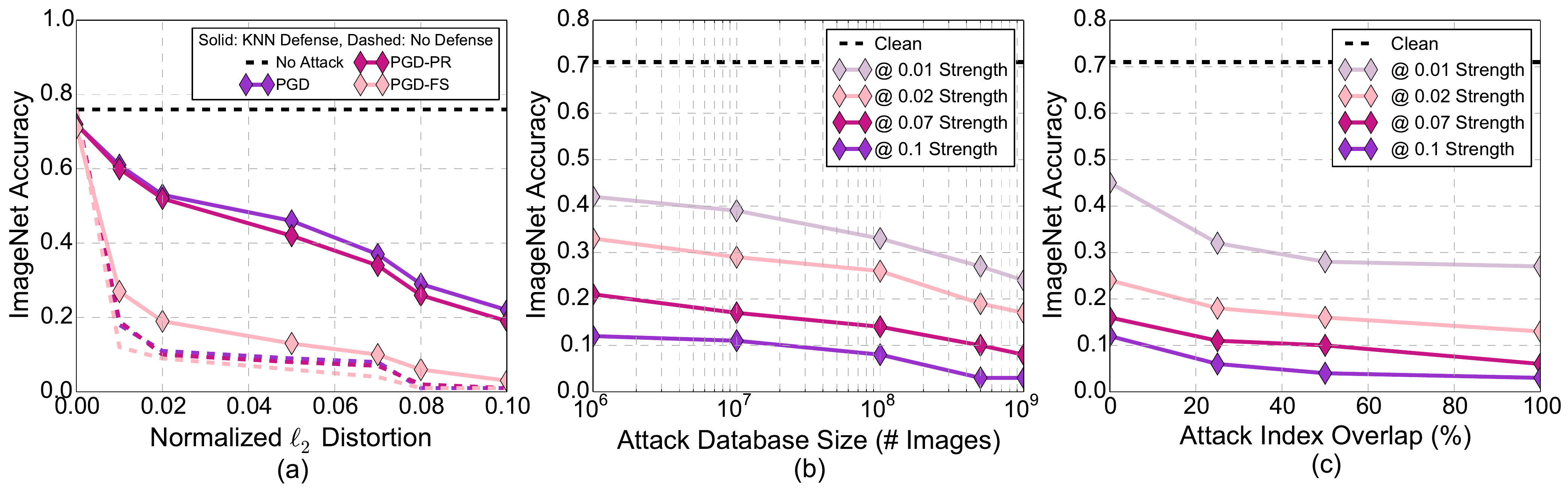}
\caption{(a) Comparison of regular PGD, PGD-PR and PGD-FS on ResNet-50 in the \textit{white-box} setting, i.e., attacker has access to the same database as the KNN defense (in this case, it is IG-1B-Targeted), (b) Variation of classification performance under KNN-PGD attack on ResNet-50 with IG-500M-Targeted defense database as the attacker database size increases, (c) Variation of classification performance under KNN-PGD attack on ResNet-50 using IG-500M-Targeted as defense database, as the overlap of images between the attacker database and the defense database increases. Note that in (c) the attack database size is always 500M.}
\label{fig:attack_white_black}
\end{figure*}
\subsection{Experiments}
We perform experiments in which we evaluate the effectiveness of the PGD, the PGD-PR, and the PGD-FS attack. In all attacks, we follow~\cite{kannan2018adversarial} and set the number of attack iterations to $10$, we set $K\!=\!50$, and use \texttt{conv\_5\_1} features to implement $g(\cdot)$. 

We assume a pure white-box setting in which the adversary uses the defense image database (IG-1B-Targeted) as its attack database, and has access to both the ResNet-50 classification model and the feature-generating function $g(\cdot)$. Figure~\ref{fig:attack_white_black}(a) shows the accuracy of a ResNet-50 using the CBW-D defense as a function of the attack strength for all three attacks. The results show that in this pure white-box setting, CBW-D only provides limited robustness against defense-aware attacks.

%{\bf{Gray-Box Attack: }}We keep both the defense and attack models as ResNet-50 but assume that attacker does not have full information about the index. We use IG-500M-Targeted as the defense index and consider the following settings:
We perform ablation experiments for the PGD-FS attack to investigate the effect of: (1) the size of the attack image database and (2) the amount of overlap between the attack and defense databases. These ablation experiments emulate a gray-box attack scenario in which the attacker only has partial access to the image database used by the defender.

\textbf{Effect of index size.} We construct the attack databases of varying size by randomly selecting a subset from the 500M images in the defense index, and measure the accuracy of our models under PGD-FS attacks using these attack databases. Figure~\ref{fig:attack_white_black}(b) displays accuracy as a function of attack database size, showing that the performance of our defense degrades as the size of attack database increases.

\textbf{Effect of attack and defense index overlap.} We fix the size of the defense and attack databases to 500M images, but we vary the percentage of images that is present in both databases. Figure~\ref{fig:attack_white_black}(c) shows the accuracy of our models under PGD-FS attacks as a function of the intersection percentage. We observe that the accuracy of our defense degrades rapidly as the overlap between the attack and the defense database grows. However, our nearest-neighbor defense is effective when in the overlap between both databases is limited.

The experimental results show that nearest-neighbor defense strategies can be effective even when the adversary is aware of the defense strategy and adapts its attack accordingly, provided that the defender can perform some kind of ``data obfuscation''. Such security-by-obfuscation is insecure in true white-box attack scenarios, but in real-world attack settings, it may be be practical because it is difficult for the adversary to obtain the same set of hundreds of millions of images that the defender has access to. %As a result, web-scale nearest neighbor may be a practical defense in gray-box attack scenarios.

%
% \begin{figure*}[t]
% \centering
% \small
% \includegraphics[width=0.9\linewidth]{figures/knnattack_combined.pdf}
% \caption{Adversarial robustness of ResNet-50 under the defense-aware PGD attacks in the gray-box settings. Figure (a)  demonstrates the robustness against an attack with a smaller disjoint index, Figure (b) demonstrates the robustness against an attack with smaller subset index and Figure (c) demonstrates the robustness against an attack with an identically sized disjoint index. In all cases, the PGD-Regular accuracy denotes the indicative strength of a regular PGD attack (without an index), and we can see that for larger indexes, the feature-based PGD attacks are stronger.}
% \label{fig:knn_attack}
% \end{figure*}

\section{Discussion and Future Work}
In this study, we have explored the feasibility of using defense strategies web-scale nearest-neighbor search to provide image-classification systems robustness against adversarial attacks. Our experiments show that, whilst such defenses are not effective in pure white-box attack scenarios, they do perform competitively in realistic gray-box settings in which the adversary knows the defense strategy used but cannot access the exact web-scale image database.

Qualitative analyses of the nearest neighbors of adversarial images show that the model predictions for these neighbors images are generally closely related to the true label of the adversarial image, even if the prediction is incorrect. On ImageNet, the prediction errors are often ``fine-grained''~\cite{anish2017synth}: for example, changing the label {\textit{golden retriever}} into {\textit{labrador}}. As a result, we believe that studies such as ours tend to overestimate the success rate of the adversary: if, for example, an attacker aims to perturb an objectionable image so that it can pass as benign, it often cannot get away with fine-grained changes to the model prediction. Randomly choosing a target class has been explored previously~\cite{anish2017synth} and is a more realistic attack setting, since this attack setting is more likely to perturb predictions to a class that is semantically unrelated.

%As long as the prediction changes from one fine-grained objectionable label to another, the system will still flag the image. Making both defense as well as attack methods {\textit{semantics-aware}} can be a fruitful direction to explore.

Our results provide the following avenues for future research. Specifically, they suggest that varying the depth of the network that constructs features for the nearest-neighbor search trades off adversarial robustness for accuracy on clean images: ``early'' features are more robust to attacks but work less well on clean images. This suggests future work should explore combinations of either features or nearest neighbors~\cite{papernot2018deepk} at different depths. Future work should also investigate approaches that decrease the similarity between the feature-producing model and the image-classification model used in our approach, as well as train these networks using adversarial-training approaches such as adversarial logit pairing~\cite{kannan2018adversarial}. Other directions for future work include developing better strategies for selecting images to use in the image database, \emph{e.g.}, using supervision from hashtags or text queries associated with the images.

\textbf{Acknowledgements.} We thank Matthijs Douze, Jeff Johnson, Viswanath Sivakumar, Herv{\'e} Jegou, and Jake Zhao for many helpful discussions and code support, and Anish Athalye and Shibani Santurkar for their comments on an earlier draft of this paper.

{\small
\bibliographystyle{ieee}
\bibliography{egbib}
}
\clearpage
\section*{Appendix A: Adversarial Attack Methods}
\textbf{Fast Gradient Sign Method (FGSM)}~\cite{goodfellow2014explaining} is one of the earliest attack techniques that has been demonstrated to successfully produce adversarial samples. The FGSM attack produces adversarial samples using the update rule:
\begin{equation*}
        \x^*_{\textsf{FGSM}} = \x + \varepsilon \cdot \sign (\nabla_{\x} L(\x, y)),
\end{equation*}
where $\x$ is the unperturbed input. When the model is available to the attacker (\textit{white-box} setting), the attack can be run using the true gradient $\nabla_\x  L(h(\x), y)$, however, in \textit{gray-box} and \textit{black-box} settings, the attacker may habe access to a surrogate gradient $\nabla_\x  L(h'(\x), y)$, which in practice, has been shown to be effective as well. A stronger version of this attack is the \textbf{Iterative Fast Gradient Sign Method (I-FGSM)}~\cite{kurakin2016adversarial}, where the adversarial input is generated by iteratively applying the FGS update over $m = \{1, ..., M\}$ steps, following:
\begin{multline*}
        \x^{(m)} = \x^{(m-1)} + \varepsilon \cdot \sign (\nabla_{\x^{(m-1)}} L(\x^{(m-1)}, y), \\ \text{where } \x^*_{\textsf{IFGSM}} = \x^{(M)}, \ \text{and } \x^{(0)} = \x
\end{multline*}
A general version of the I-FGSM attack is the \textbf{Projected Gradient Descent (PGD)}~\cite{madry2017towards} attack, which clips the gradients to project them back to the feasible image domain, and also includes random restarts in the optimization process. We use this attack over the I-FGSM attack due to its stronger nature. The FGSM, I-FGSM, and PGD attacks approximately minimize the Chebyshev distance between the input $\x$ and the generated adversarial sample $\x^*$. \textbf{Carlini-Wagner's $\ell_p$ (CW-Lp)} attack attempts to find a solution to an unconstrained optimization problem that jointly penalizes a differentiable surrogate for the model accuracy along with a distance measure for regularization, such as the $\ell_2$ or $\ell_\infty$ distance.
\begin{multline*}
        \x^*_{\textsf{CW-Lp}} = \min_{\x'}\big[\lVert \x - \x' \rVert_p^2 +  \lambda_f \max(-\kappa, Z(\x')_{h(\x)} \\ - \max\{ Z(\x')_k : k \neq h(\x)\}  )\big]
\end{multline*}
%Similarly, the \textbf{Carlini-Wagner's $\ell_\infty$ (CWLINF)} attack obtains adversarial samples by solving:
%\begin{equation*}
%       \x^*_{\textsf{CWLINF}} = \min_{\x'}\big[\lVert \x - \x' \rVert_\infty^2 + \lambda_f \max(-\kappa, Z(\x')_{h(\x)} - \max\{ Z(\x')_k : k \neq h(\x)\}  )\big]
%\end{equation*}
Herein, $\kappa$ denotes a margin parameter, and the parameter $\lambda_f$ relatively weighs the losses from the distance penalty and accuracy surrogate (hinge loss of predicting an incorrect class). The most common values for $p$ are $p\!=\!2$ and $p\!=\!\infty$; we use the implementation of~\cite{guo2017countering} to implement FGSM, IFGSM, PGD, and CW-L2. For all the above attacks, we enforce that the image remains within $[0,1]^d$ by clipping values.

\section*{Appendix B: Feature Construction}
% \subsection{Choice of Feature Extractor ($\phi$)}
\label{sec:featextractor}
To evaluate the trade-off between robustness and accuracy of neural network features at different depths (layers), we use features extracted from different layers of ResNet-50 models~\cite{he2016deep} as the basis for retrieving nearest neighbors. Since layers are very high-dimensional, we reduce the final dimensionality of the feature vectors to 256 by performing a spatial average pooling followed by PCA (see Table~\ref{tab:features} for a complete description). We found that spatial average pooling step helps in increasing accuracy as well as the computational efficiency of PCA.

\begin{table}[h]
\centering
\begin{tabular}{l|cc}
\hline
\hline
\textbf{Feature Layer} & \textbf{Uncompressed Size} & \textbf{Pooled Size}\\ \hline
\texttt{conv\_2\_3} & $256\times56\times56$ & $256\times7\times7$ \\ \hline
\texttt{conv\_3\_4} & $ 512\times 28 \times 28$ & $512\times 7 \times 7$ \\ \hline
\texttt{conv\_4\_6} & $1024\times 14 \times 14$ & $1024\times 4 \times 4$ \\ \hline
\texttt{conv\_5\_1} & $2048\times 7 \times 7$ & $2048\times 1 \times 1$ \\
\hline\hline
\end{tabular}
\caption{Feature vector details. All features are finally compressed to a dimensionality of 256 by a PCA done over 3M samples.}
\label{tab:features}
\end{table}

The features are extracted using the PyTorch~\cite{paszke2017automatic} framework; we use the SciPy~\cite{scipy} implemention of online PCA for dimensionality reduction. The PCA was computed on 3 million randomly selected samplesfor the {\it IG} and {\it YFCC} datasets, and on the complete training set for ImageNet.

For nearest-neighbor matching, we construct a pipeline using the GPU implementation of the FAISS~\cite{johnson2017billion} library; we refer the readers to the original paper for more details about billion-scale fast similarity search.

\section*{Appendix C: Hard versus Soft\\~~~Combination of Predictions}
\begin{table*}[ht!]
\centering
\setlength\tabcolsep{2pt}
\resizebox{\textwidth}{!}{
\begin{tabular}{l|ccc|ccc|ccc|ccc|ccc|ccc}
\toprule
\multirow{3}{*}[-1.2em]{\textbf{Image database}} & \multicolumn{6}{c|}{\textbf{Clean}} & \multicolumn{6}{c|}{\textbf{Gray box}} & \multicolumn{6}{c}{\textbf{Black box}} \\
& \multicolumn{3}{c|}{\underline{Soft Combination}} & \multicolumn{3}{c|}{\underline{Hard Combination}} & \multicolumn{3}{c|}{\underline{Soft Combination}} & \multicolumn{3}{c|}{\underline{Hard Combination}} & \multicolumn{3}{c|}{\underline{Soft Combination}} & \multicolumn{3}{c}{\underline{Hard Combination}} \\
& UW & CBW-E & CBW-D & UW & CBW-E & CBW-D & UW & CBW-E & CBW-D & UW & CBW-E & CBW-D & UW & CBW-E & CBW-D & UW & CBW-E & CBW-D \\ \midrule
IG-50B-All (\texttt{conv\_5\_1-RMAC}) & 0.632 & 0.644 & {\bf 0.676} & 0.637 & 0.642 & 0.649 & 0.395 & 0.411 & {\bf 0.427} & 0.402 & 0.403 & 0.414 & 0.448  & 0.459 & {\bf 0.491} & 0.457 & 0.462 & 0.473 \\
IG-1B-Targeted (\texttt{conv\_5\_1}) & 0.659 & 0.664 & {\bf 0.681} & 0.668 & 0.671 & 0.673 & 0.415 & 0.429 & {\bf 0.462} & 0.418 & 0.423 & 0.437 & 0.568 & 0.574 & {\bf 0.587} & 0.554 & 0.561 & 0.571\\
IN-1.3M (\texttt{conv\_5\_1}) & 0.472 & 0.469 & 0.471 & {\bf 0.475} & 0.472 & 0.473 & 0.285 & 0.286 & 0.286 & 0.291 & 0.289 & {\bf 0.293} & 0.311 & 0.312 & 0.312 & {\bf 0.316} & 0.309 & 0.314 \\
\bottomrule
\end{tabular}
}
\caption{ImageNet classification accuracies of ResNet-50 on PGD-generated images with a normalized $\ell_2$ distance of $0.06$, using our nearest-neighbor defenses with three different image databases on both soft and hard combination techniques, with three different weighing strategies (UW, CBW-E and CBW-D), and $K\!=\!50$. Accuracies on clean images are included for reference.}
\label{tab:apvscbv}
\end{table*}
As discussed in the main paper, we evaluate both hard and soft prediction combinations, as described below:

\textbf{Soft Combination (SC)}: We return the \textit{weighted average} of the softmax probability vector of all the nearest neighbors. The predicted class is then the $\arg\max$ of this average vector. This corresponds to a ``soft'' combination of the model predictions over the data manifold.\vspace*{0.05in}

\textbf{Hard Combination (HC):} In this case, each nearest neighbor votes for its ``hard'' predicted class with some weight. The final prediction is then taken as the most commonly predicted class. %If the classification model $h$ would achieve zero classification error on the image database (\emph{i.e.}, if $h$ would classify all images in the database correctly), then this strategy approaches the Bayes-optimal classifier \cite{wang2017analyzing}.

In Table~\ref{tab:apvscbv}, we present the results of both these approaches with uniform weighing (UW) and confidence-based weighing (CBW). These results were obtained using the same experimental setup as described in the main paper.

\section*{Appendix D: Results with FGSM and CWL-2}
Below, we present results that measure the effectiveness of our defense strategy against the FGSM and CWL-2 attacks. All experiments follow the same experimental setup as described in the main paper (but use a different attack method).

\begin{figure}[ht]
\centering
\small
\centering
\includegraphics[width=0.8\linewidth]{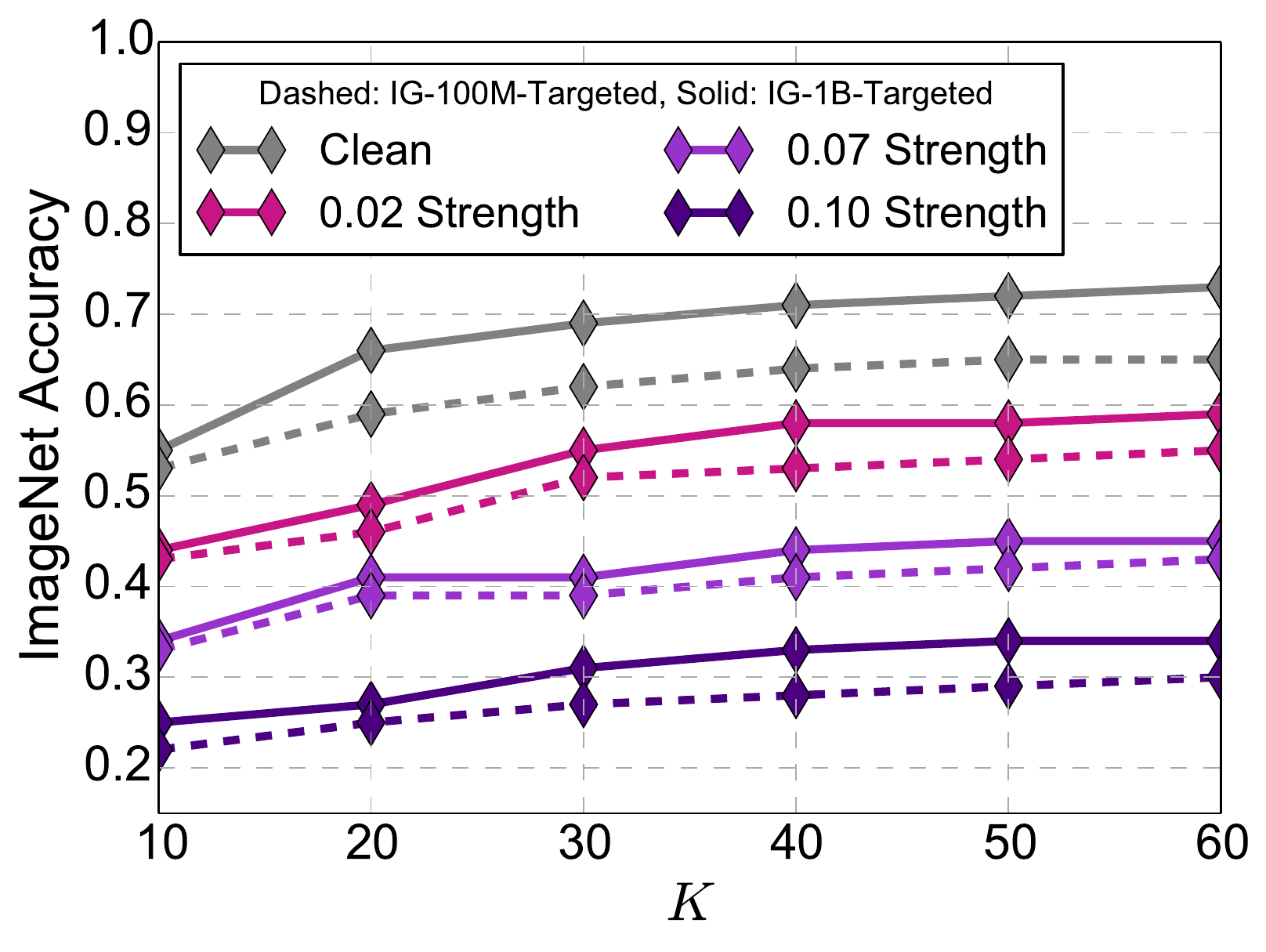}
\caption{Classification accuracy of ResNet-50 using our CBW-D defense on FGSM adversarial ImageNet images, as a function of the normalized $\ell_2$ norm of the adversarial perturbation. Defenses are implemented via nearest-neighbor search using \texttt{conv\_5\_1} features on the IG-1B-Targeted (solid lines) and IG-100M-Targeted (dashed lines). Results are for the black-box setting.}
\label{fig:k_variation_fgsm}
\end{figure}

\begin{figure}[ht]
\centering
\small
\centering
\includegraphics[width=0.8\linewidth]{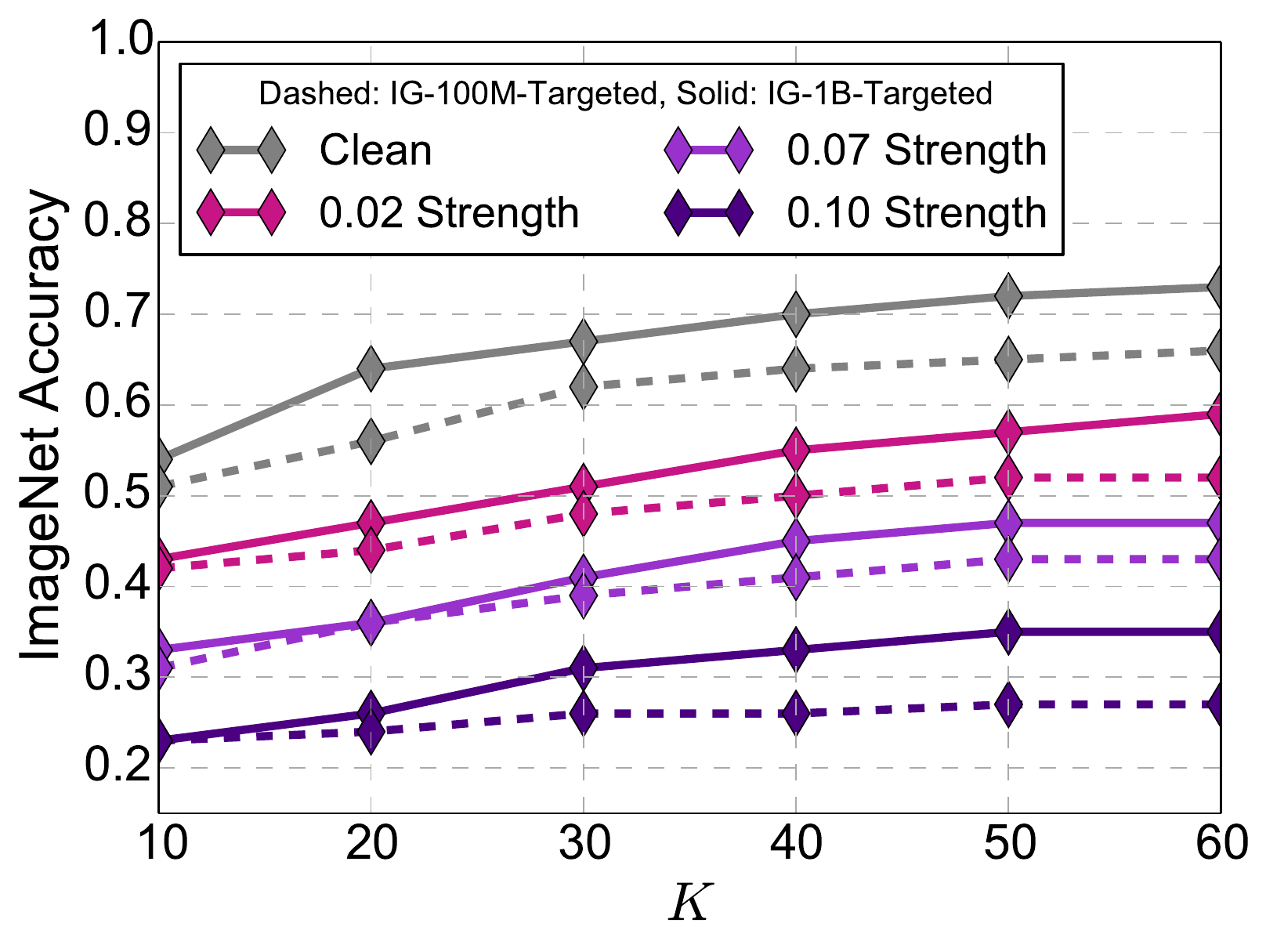}
\caption{Classification accuracy of ResNet-50 using our CBW-D defense on CWL-2 adversarial ImageNet images, as a function of the normalized $\ell_2$ norm of the adversarial perturbation. Defenses are implemented via nearest-neighbor search using \texttt{conv\_5\_1} features on the IG-1B-Targeted (solid lines) and IG-100M-Targeted (dashed lines). Results are for the black-box setting.}
\label{fig:k_variation_cwl2}
\end{figure}

\paragraph{Effect of Number of Neighbors, $K$.} Figures~\ref{fig:k_variation_fgsm} and~\ref{fig:k_variation_cwl2} describe the effect of varying $K$ under FGSM and CWL-2 attacks. We observe similar trends as for the PGD attacks.

\begin{figure*}[ht]
\centering
\small
\includegraphics[width=0.9\linewidth]{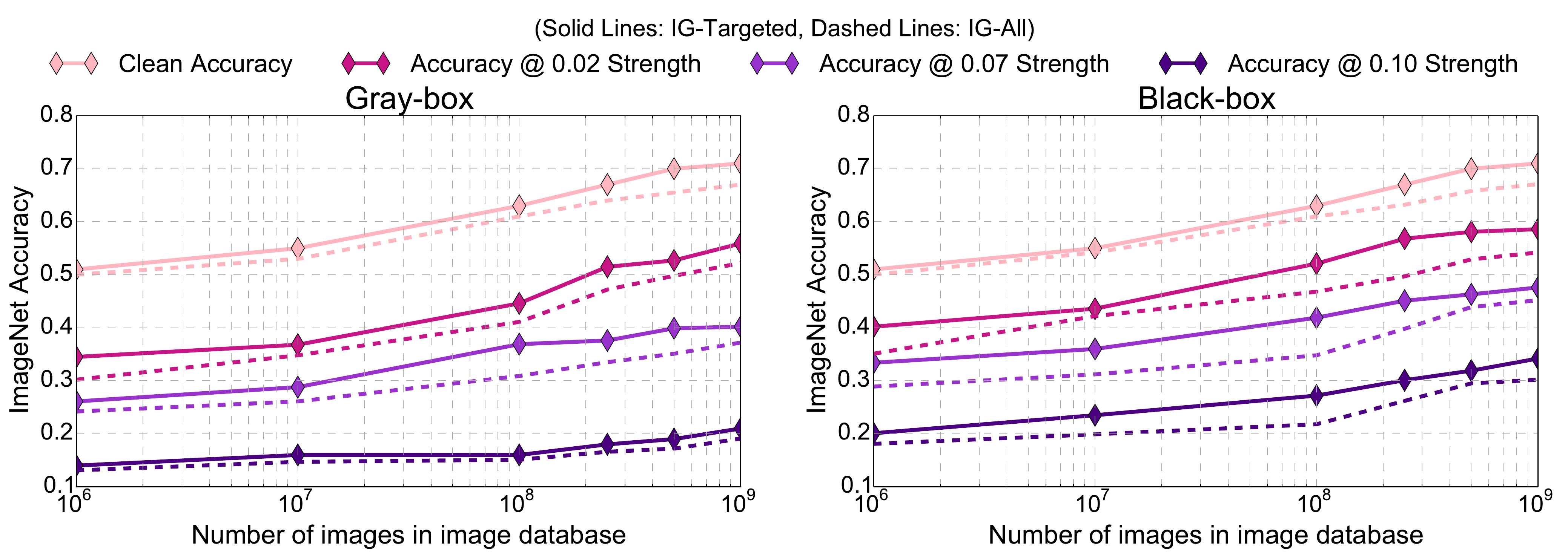}
\caption{Classification accuracy of ResNet-50 using the CBW-D defense on FGSM adversarial ImageNet images, using the IG-$N$-Targeted database (solid lines) and IG-$N$-All database (dashed lines) with different values of $N$. Results are presented in the gray-box (left) and black-box (right) settings.}
\label{fig:size_variation_fgsm}
\end{figure*}

\begin{figure*}[ht]
\centering
\small
\includegraphics[width=0.9\linewidth]{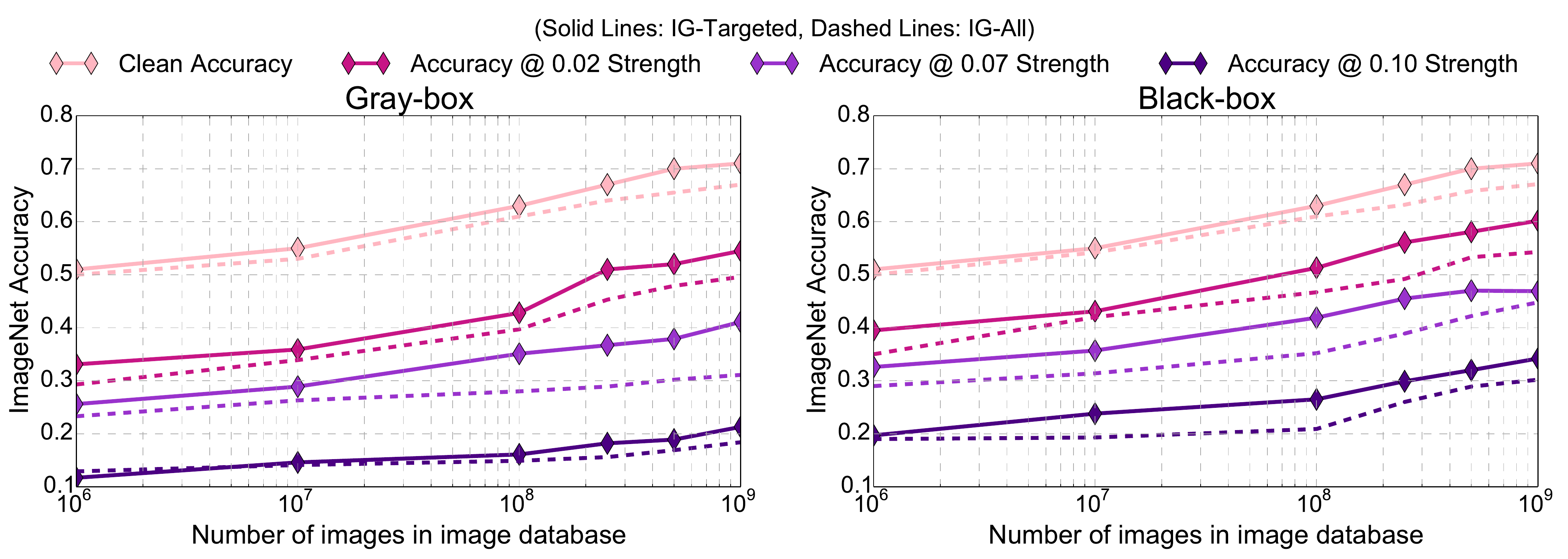}
\caption{Classification accuracy of ResNet-50 using the CBW-D defense on CWL-2 adversarial ImageNet images, using the IG-$N$-Targeted database (solid lines) and IG-$N$-All database (dashed lines) with different values of $N$. Results are presented in the gray-box (left) and black-box (right) settings.}
\label{fig:size_variation_cwl2}
\end{figure*}

\paragraph{Effect of Image Database Size.} Figure~\ref{fig:size_variation_fgsm} and~\ref{fig:size_variation_cwl2} describe the effect of varying the index size under FGSM and CWL-2 attacks. We observe similar trends as for the PGD attacks.

\begin{figure*}[ht]
\centering
\small
\centering
\includegraphics[width=0.9\linewidth]{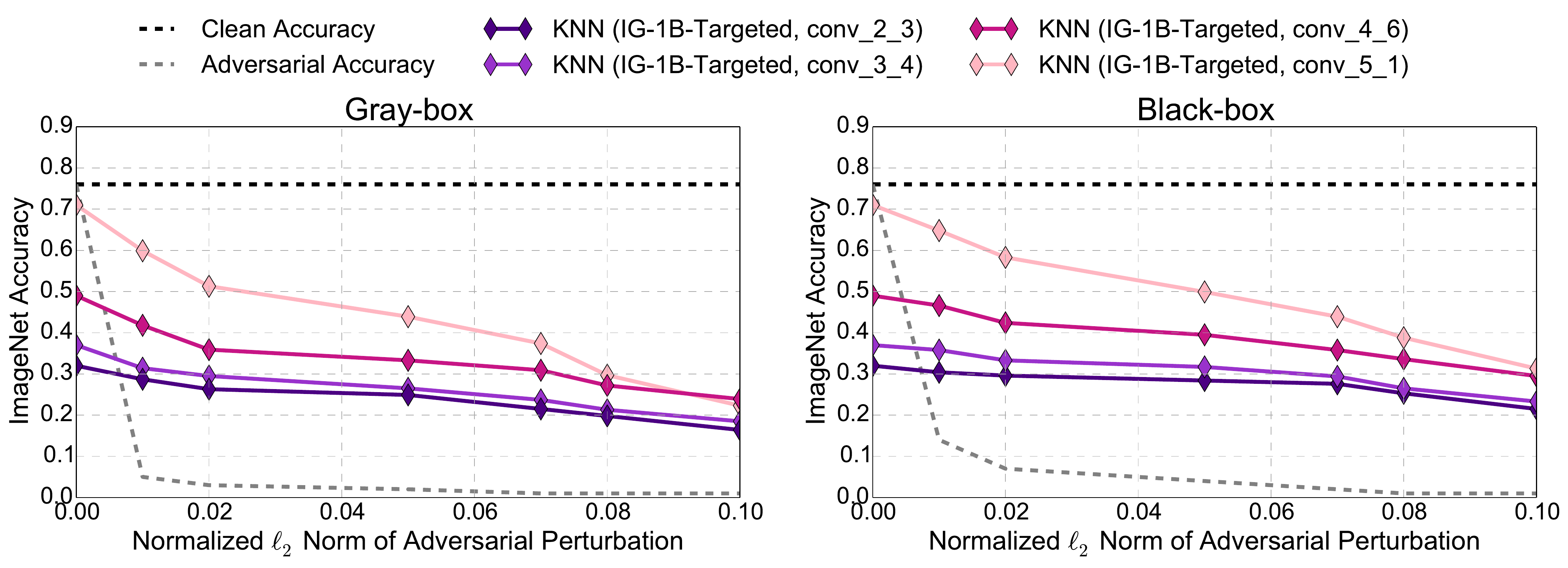}
\caption{Classification accuracy of ResNet-50 using the CBW-D defense on FGSM adversarial ImageNet images, as a function of the normalized $\ell_2$ norm of the adversarial perturbation. Defenses use four different feature representations of the images in the IG-1B-Targeted image database. Results are presented for the gray-box (left) and black-box (right) settings.}
\label{fig:feature_variation_fgsm}
\end{figure*}

\begin{figure*}[ht]
\centering
\small
\centering
\includegraphics[width=0.9\linewidth]{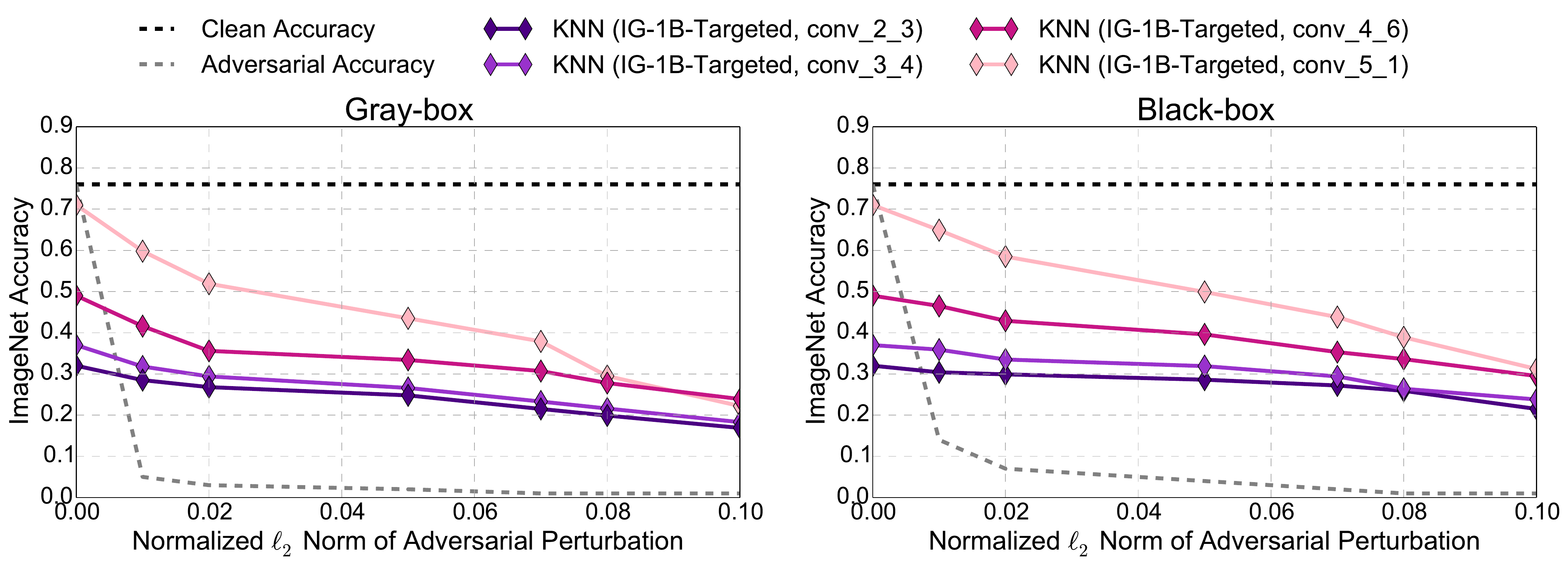}
\caption{Classification accuracy of ResNet-50 using the CBW-D defense on CWL-2 adversarial ImageNet images, as a function of the normalized $\ell_2$ norm of the adversarial perturbation. Defenses use four different feature representations of the images in the IG-1B-Targeted image database. Results are presented for the gray-box (left) and black-box (right) settings.}
\label{fig:feature_variation_cwl2}
\end{figure*}

\paragraph{Effect of Feature Space.} Figure~\ref{fig:feature_variation_fgsm} and~\ref{fig:feature_variation_cwl2} present classification accuracies obtained using CBW-D defenses based on four different feature representations of the images in the IG-1B-Targeted database, for the FGSM and CWL-2 attacks respectively. We observe a similar trend as in the case of PGD attacks.

% {\small
% \bibliographystyle{ieee}
% \bibliography{egbib}
% }

%\input{appendixa}
\end{document}